\pdfoutput=1

\documentclass[11pt]{article}

\usepackage[preprint]{acl}

\usepackage{times}
\usepackage{latexsym}

\usepackage[T1]{fontenc}

\usepackage[utf8]{inputenc}

\usepackage{microtype}

\usepackage{inconsolata}

\usepackage{graphicx}

\usepackage{multirow}
\usepackage{arydshln} 
\usepackage{booktabs} 
\usepackage{caption} 

\usepackage{natbib} 
\usepackage{hyperref} 

\usepackage{enumitem}

\title{Can Vision-Language Models Infer Speaker’s Ignorance?\\
The Role of Visual and Linguistic Cues}

\author{Ye-eun Cho \\
  English language and literature\\
  Sungkyunkwan University\\
  Seoul, South Korea\\
  \texttt{joyenn@skku.edu} \\\And
  Yunho Maeng \\
  Ewha Womans University \&\\
  LLM Experimental Lab, MODULABS \\
  Seoul, South Korea\\
  \texttt{yunhomaeng@ewha.ac.kr} \\}

\begin{document}
\maketitle
\begin{abstract}
This study investigates whether vision language models (VLM) can perform pragmatic inference, focusing on ignorance implicatures, utterances that imply the speaker’s lack of precise knowledge. To test this, we systematically manipulated contextual cues: the visually depicted situation (visual cue) and QUD-based linguistic prompts (linguistic cue). When only visual cues were provided, three state-of-the-art VLMs (GPT-4o, Gemini 1.5 Pro, and Claude 3.5 sonnet) produced interpretations largely based on the lexical meaning of the modified numerals. When linguistic cues were added to enhance contextual informativeness, Claude exhibited more human-like inference by integrating both types of contextual cues. In contrast, GPT and Gemini favored precise, literal interpretations. Although the influence of contextual cues increased, they treated each contextual cue independently and aligned them with semantic features rather than engaging in context-driven reasoning. These findings suggest that although the models differ in how they handle contextual cues, Claude’s ability to combine multiple cues may signal emerging pragmatic competence in multimodal models.
\end{abstract}

\section{Introduction}
In recent years, many large language models (LLMs) have demonstrated the ability to solve a wide variety of tasks, contributing to their growing popularity. Initially limited to text-based inputs, these models have been extended to incorporate visual inputs, paving the way for vision-language models (VLMs). By bridging vision and language modalities, VLMs have expanded the possibilities for AI applications and become central to the ongoing technological revolution \citep{radford2021learning, ramesh2021zero, alayrac2022flamingo, li2023blip}.

VLMs have enabled various multimodal applications, such as object recognition \citep{ren2015faster, chen2020simple, he2020momentum}, caption generation \citep{vinyals2015show, chen2022pali, yu2022coca}, and visual question answering \citep{antol2015vqa}. These tasks primarily focus on associations between visual and textual inputs by identifying objects, describing scenes, or responding to straightforward queries. While such capabilities are remarkable, they represent only the surface level of human-like understanding. In fact, real-world communication often requires reasoning about implicit meanings that emerge from the interplay between language and visual information \citep{sikka2019deep}. To move toward more human-like multimodal intelligence, VLMs must also be able to engage in this type of context-sensitive and inferential processing (see \citealp{kruk2019integrating}). This raises critical questions about VLMs’ capacity for context-sensitive reasoning, which underlies the pragmatic competence required for real-world communication.

Pragmatics offers an ideal framework for investigating this question. In human communication, pragmatic inference plays a crucial role in understanding intended meanings beyond literal statement \citep{grice75, wilson95, levinson2000presumptive}. Contextual cues often provide disambiguating information that influences the interpretation of utterances, making pragmatic reasoning inherently multimodal \citep{clark1996using, kendon2004gesture, martin2007multimodal, mcneill2008gesture}. While a few studies on pragmatic reasoning have been explored in text-only LLMs \citep{hu2022predicting, hu2023expectations, lipkin2023evaluating, cho2024pragmatic, capuano2024pragmatic, tsvilodub2024experimental}, the visual modality enriches meaning construction through interaction with linguistic input. As visual context provides rich, implicit information that influences language interpretation, studying pragmatic phenomena through VLMs presents an intriguing research opportunity. However, how well VLMs can leverage visual information for pragmatic inference remains largely unexplored.

Therefore, we investigate whether VLMs exhibit sensitivity to context, particularly focusing on ignorance implicatures—a pragmatic phenomenon in which a speaker’s utterance implies a lack of precise knowledge, and whether this sensitivity can be modulated by a single cue or by the combination of multiple cues. By examining how VLMs handle this phenomenon in comparison to human reasoning, we aim to better understand their strengths and limitations in processing context-dependent pragmatic meaning.

\section{Ignorance implicatures}

\setlength{\tabcolsep}{5pt} 
\renewcommand{\arraystretch}{1.9} 
\begin{table*}
  \centering
    \resizebox{\textwidth}{!}{%
    \begin{tabular}{|c|c|p{7cm}|c|c|}
    \hline
    \rule{0pt}{1pt}
        \textbf{No.} & \textbf{Image} & \centering \textbf{Text} & \textbf{Situation} & \textbf{Modifier} \\
    \hline
    1 & \multirow{3}{*}{\includegraphics[width=0.3\textwidth, height=0.3\textwidth, keepaspectratio]{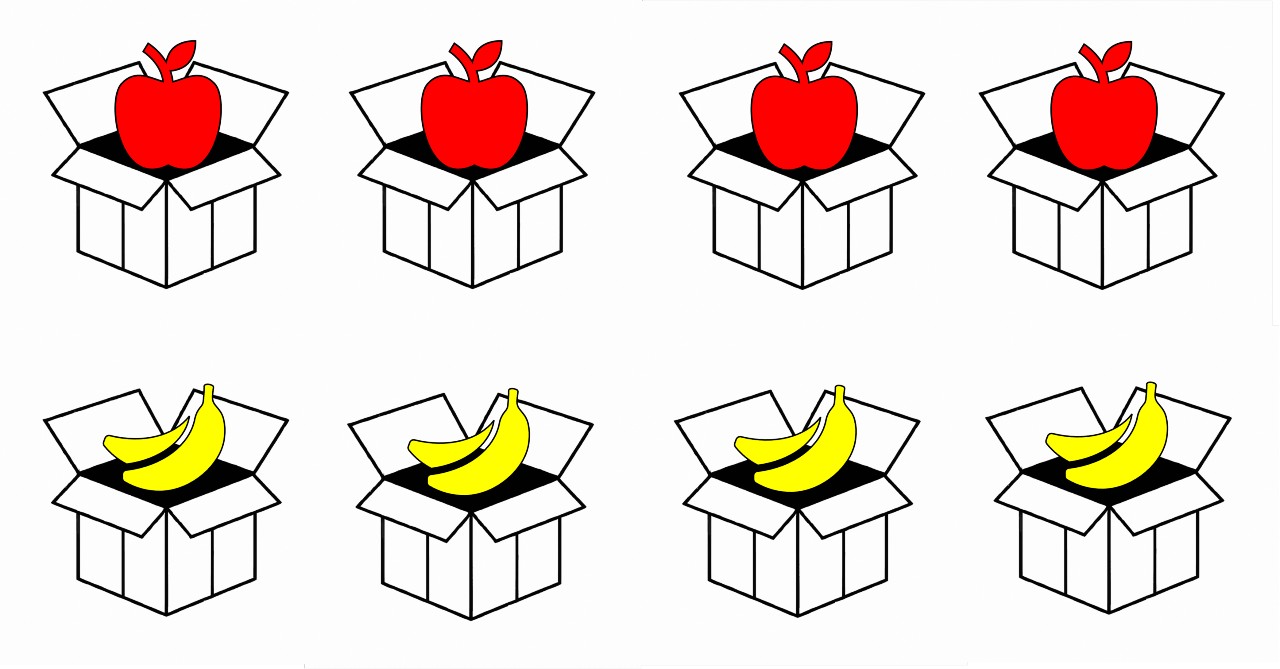}} & \centering \textit{There are four apples in the boxes.} & \centering precise & \centering bare \tabularnewline
    \cline{3-5}
       & & \centering \textit{There are at least four apples in the boxes.} & \centering precise & \centering superlative \tabularnewline
    \cline{3-5}
       & & \centering \textit{There are more than three apples in the boxes.} & \centering precise & \centering comparative \tabularnewline
    \hline
    2 & \multirow{3}{*}{\includegraphics[width=0.3\textwidth, height=0.3\textwidth, keepaspectratio]{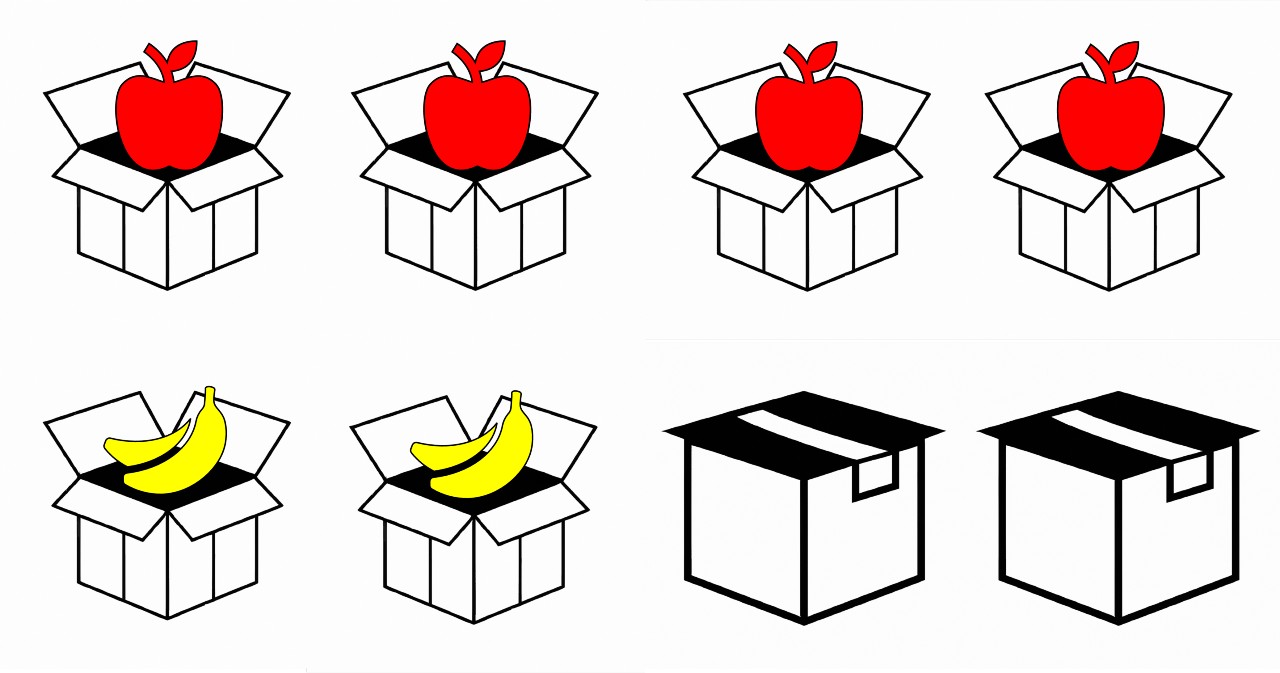}} & \centering \textit{There are four apples in the boxes.} & \centering approximate & \centering bare \tabularnewline
    \cline{3-5}
       & & \centering \textit{There are at least four apples in the boxes.} & \centering approximate & \centering superlative \tabularnewline
    \cline{3-5}
       & & \centering \textit{There are more than three apples in the boxes.} & \centering approximate & \centering comparative \tabularnewline
    \hline
    \end{tabular}
    }
  \caption{A sample set of experimental materials}
\end{table*}

To better understand pragmatic reasoning in real-world language use, we examine the phenomenon of \textit{ignorance implicatures}. Consider the examples in (1).

\vspace{1em}  
\noindent
(1)
\begin{description}[labelindent=1em, leftmargin=2.5em, style=nextline, itemsep=0pt, parsep=0pt, topsep=2pt, font=\normalfont]
  \item[\textnormal{a.} \textit{(bare numeral)}]\hspace*{0em}Four students passed the exam.
  \item[\textnormal{b.} \textit{(superlative modified numeral)}]\hspace*{0em}At least four students passed the exam.
  \item[\textnormal{c.} \textit{(comparative modified numeral)}]\hspace*{0em}More than three students passed the exam.
\end{description}
\vspace{1em}

When it comes to how many students passed the exam, the statement (1a) triggers ‘exactly four’ interpretation, whereas (1b) and (1c) do not. Both (1b) and (1c) contain modified numerals, suggesting that the speaker may not know the exact number of students who passed. This is known as ignorance implicatures, where the speaker’s choice of modifier implies a lack of precise knowledge.

However, not all modifiers give rise to ignorance implicatures to the same extent. Previous studies have shown that superlative modifiers like at least tend to trigger ignorance implicatures more consistently than comparative ones like more than \citep{nouwen2010two, cummins2012granularity, coppock2013raising, mayr2014more, cremers2022ignorance}. In this regard, the likelihood of ignorance inferences typically follows the hierarchy: \textit{superlative modified numerals > comparative modified numerals > bare numerals}.

This observation has prompted researchers to explore how such inferences arise, leading to two main perspectives. One approach suggests that ignorance inference is dependent on the words or phrases themselves (\citealp{geurts2007least, nouwen2010two}; also see \citealp{geurts2010scalar}). \citet{geurts2007least}, for example, argued that the semantics of superlative modifiers are inherently more complex. While \textit{more than n} expresses a simple meaning ‘larger than \textit{n}’, \textit{at least n} can convey both ‘possible that there is a set of \textit{n}’ and ‘certain that there is no smaller set of \textit{n}.’ According to \citet{nouwen2010two}, when someone has basic knowledge of geometry, (2a) gives the impression that the speaker lacks precise information, as compared to (2b). This attributes the ignorance implicatures to a semantic property specific to \textit{at least}. 

\vspace{1em}
\noindent
\begin{tabular}{@{}ll}
(2) &
\begin{minipage}[t]{0.85\textwidth}
a. ? A hexagon has at least five sides. \\
b. A hexagon has more than four sides.
\end{minipage}
\end{tabular}
\vspace{1em}

Under the pragmatic account, on the other hand, ignorance implicatures for both at least and more than have been primarily explained through Gricean reasoning, particularly the Maxim of Quantity \citep{grice75}, which holds that the speaker’s choice to provide a lower-bound statement, rather than a more informative exact number, suggests that the speaker lacks precise knowledge \citep{buring2008least, cummins2010comparative, coppock2013raising}. More recent studies have expanded this account by emphasizing the role of contextual factors \citep{cummins2012granularity, cummins2013modelling, mayr2014more, westera2014ignorance, cremers2022ignorance}. In these studies, contextual cues, including Question Under Discussion (QUD), preceding discourse, or accompanying visual input, were manipulated to modulate the likelihood of implicature.

For instance, \citet{westera2014ignorance} investigated how different types of modified numerals give rise to ignorance implicatures depending on contextual demands and processing cost. In their experiments, participants read short dialogues or utterances containing modified numerals and judged how confident the speaker seemed about the exact quantity, as well as how natural the utterance was. To manipulate the informativeness required by the discourse, the authors introduced different QUDs, such as a ‘how many’ condition (\textit{How many of the diamonds did you find under the bed?}), which demanded precise answers, and a ‘polar’ condition (\textit{Did you find \{at most | less than\} ten of the diamonds under the bed?}), which did not require numerically specific responses, as they could be answered with a simple \textit{yes} or \textit{no}. The results showed that ignorance inferences occurred significantly more consistently when the QUD demanded precision (‘how many’ condition), suggesting that contextual expectations about informativeness directly affect how such inferences are drawn.

Likewise, \citet{cremers2022ignorance} systematically manipulated various contextual factors to investigate the conditions under which ignorance implicatures arise. Their experiments involved multiple levels of visual information, QUD types, and textual scenarios. In particular, visual information was used to represent the informativeness of the situation—for example, a precise situation in which all eight cards were face-up, and an approximate situation in which two of the eight cards remained face-down, obscuring the exact quantity. Their findings revealed that ignorance inferences were more likely when the QUD required a precise answer (‘howmany’ condition) and when the visual context left room for uncertainty (‘approximate’ condition). These results highlight that ignorance implicatures are largely influenced by both linguistic and non-linguistic contextual cues.

These findings raise the question of whether and how visual and linguistic contextual information can enhance the pragmatic competence of VLMs. To address this question, the present study examines whether VLMs exhibit sensitivity to ignorance implicatures across multiple contextual cues.

\section{Methods}
\subsection{Data}

\begin{figure*}[t]
  \includegraphics[width=\textwidth]{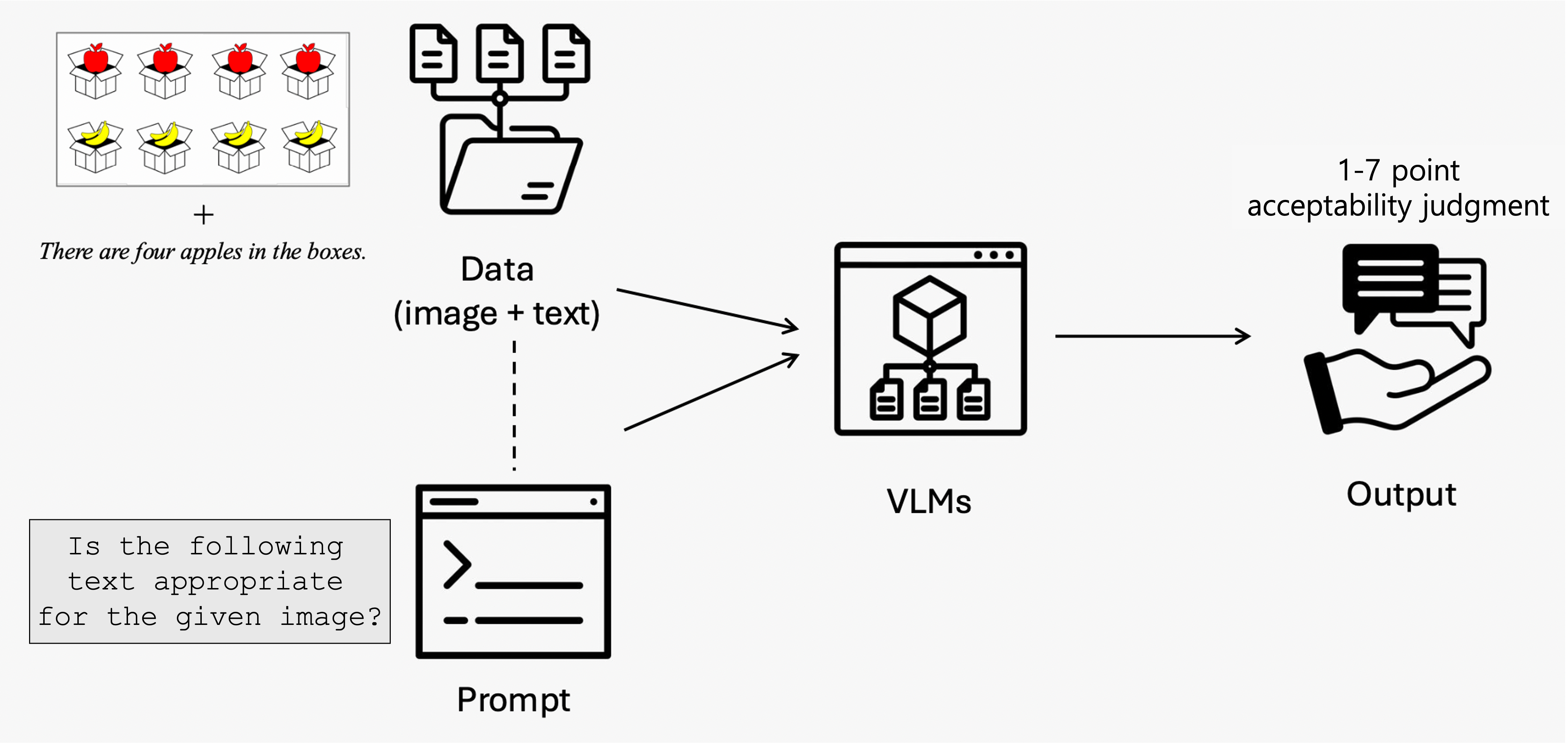}
  \caption{Overview of the experimental procedure}
\end{figure*}

As presented in Table 1, experimental materials were designed using two images for contextual precision (henceforth, ‘situation’) and texts including bare numeral, superlative, and comparative modifiers (henceforth, ‘modifier’).

In detail, images were used to manipulate the contextual precision, where a picture showing all 8 boxes open and providing the exact number of target objects was labeled as ‘precise’, and a picture with 2 out of 8 boxes remaining closed and an uncertain number of target objects was labeled as ‘approximate’. In both types of situation, the target objects consistently appeared in 4 boxes. For example, in the image for precise situation, all 8 boxes are open and 4 of them contain apples. Since all boxes are open, we can tell that the target objects are exactly 4. On the other hand, in the image for approximate situation, 2 out of the 8 boxes remain closed and 4 of the open boxes contain apples. As what is inside the closed boxes is unknown, the target objects could be 4 or more. The corresponding texts were categorized based on the modifier types, including ‘bare’ (bare numeral \textit{n}), ‘superlative’ (\textit{at least n}), and ‘comparative’ (\textit{more than n}).

Image data was created by combining open and closed boxes generated by GPT-4o \citep{openai2024gpt4o} with standard icons for target objects. In this study, the number of target objects was consistently set to four, as previous work has shown that VLMs often exhibited limited performance on numerical reasoning tasks and experience a marked decline in accuracy when counting more than four items \citep{paiss2023teaching}. Additionally, since VLMs tend to struggle with counting when objects are presented in unstructured or cluttered spatial arrangements \citep{liu2019context, rahmanzadehgervi2024vision}, the experimental images were carefully constructed with precisely aligned rows and columns. In this manner, each set of materials consisted of 2 images, each paired with 3 corresponding texts. In total, 70 sets of materials were used in the experiment.

\subsection{Models and Procedure}

As VLMs for the experiment, we used GPT-4o \citep{openai2024gpt4o}, Gemini 1.5 Pro \citep{gteam2024gemini}, and Claude 3.5 sonnet \citep{Anthropic2024claude}. These models were selected due to their ability to process both image and text inputs simultaneously. They not only allow for the matching of images with text to determine their relationship but also provide the functionality to selectively query specific parts of the text within a broader context. This makes them well-suited for a series of our experiments.

For the experiment, these models were initialized using API keys. The images were then resized to a standard size of 224x224 pixels using the Pillow library \citep{clark2015pillow} to ensure consistency in input dimensions and optimize processing efficiency. After resizing, the images were encoded into base64 format to ensure compatibility for input into the model’s API.

Each experiment involved presenting the image alongside text prompts, which were specifically tailored for each task. All the materials, code and result of the experiment are publicly available.\footnote{\url{ https://github.com/joyennn/ignorance-implicature}}

\section{Experiment 1}
\subsection{Prompt}
\citet{cremers2022ignorance} argued that the disagreement over main findings related to ignorance inference was that the detection depends on the types of tasks participants were asked to perform. Specifically, it varied depending on whether participants were given an acceptability judgment task \citep{coppock2013diagnosing, westera2014ignorance, cremers2022ignorance}, where they judged the acceptability of the given sentences with respect to the depicted scenarios or images, or an inference task \citep{geurts2010scalar}, where they judged whether \textit{exactly n} implies \textit{at least n}. \citet{cremers2022ignorance} argued that ignorance inference is more accurately assessed when evaluating the appropriateness of a sentence in relation to the context, rather than through the logical reasoning involved in inference tasks. In this regard, the acceptability judgment task serves as an effective method, guiding participants to evaluate whether a sentence is contextually appropriate. The acceptability judgment task typically involves either a true/false response format, as in truth-value judgment tasks \citep{coppock2013diagnosing}, or a numerical scale to capture the degree of ignorance implicatures in a more fine-grained manner \citep{westera2014ignorance, cremers2022ignorance}. In our experiment, we adopt a 1-7 scale to assess the appropriateness of image-text pairings in a more fine-grained manner. 

As presented in Figure 1, each model was prompted to rate whether the texts with bare numerals, superlative, and comparative modifiers are appropriate for the given image on a scale from 1 to 7. The phrases used in the prompt were adapted from Experiment 3 in \citet{cremers2022ignorance}. In this manner, 70 sets of experimental items were repeated 5 times to improve the reliability, resulting in a total of 2,100 individual responses from each of three models, as detailed below.

\vspace{1em}
\begin{tabular}{|p{0.4\textwidth}|}
\hline
Is the following text appropriate for the given image? \\
Please reply with a single integer between 1 and 7, where 1 means “not at all appropriate” and 7 means “completely appropriate.” \\
\textbf{Text:} \{text\} \\
\hline
\end{tabular}
\vspace{1em}

\begin{figure}[t]
  \includegraphics[width=\columnwidth]{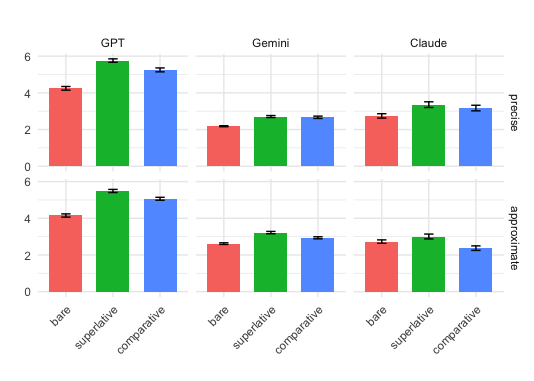}
  \caption{Result of Experiment1 — Mean scores for the appropriateness of image-text pairs based on modifiers, situations and models}
\end{figure}

\subsection{Result}

Figure 2 shows the mean scores for the appropriateness of image-text pairs based on the types of modifiers, situations, and models. For the statistical analysis, we built mixed-effects logistic regression models \citep{baayen2008analyzing, baayen2008mixed, jaeger2008categorical, jaeger2011corpus} to analyze the results for each model, using the \texttt{lme4} package \citep{bates2015package} in R software \citep{r2023}. To examine the fixed effects, modifier and situation were set as independent variables, with appropriate scores as the dependent variable. Image and text were specified as random effects. For independent variables, the bare condition and the precise condition were set as the reference levels for modifier and situation, respectively.

As a result, the scores for appropriateness of image-text pairs followed the order of \textit{superlative > comparative > bare} in both types of situations across almost all models, except for the approximate condition of Claude. While these predominant results aligned with findings from \citet{cremers2022ignorance}, where participants preferred the text with superlative and comparative in the approximate condition.  However, the similar pattern in the precise situation was unexpected. In this situation, texts containing bare numerals should have been considered more appropriate than those with the other modifiers, as the number of target objects was explicitly defined. 

Statistically, these results were influenced mostly modifier, which showed main effects in GPT (\textit{p} < 0.001), Gemini (\textit{p} < 0.001), and Claude (\textit{p} < 0.01, 0.05). However, there were no significant effects on situation alone in GPT (\textit{p} = 0.66) and Claude (\textit{p} = 0.96), nor in the interaction of situation and modifier in GPT (\textit{p} = 0.06, 0.33) and Gemini (\textit{p} = 0.11, < 0.001).

In summary, superlative and comparative modifiers, which imply uncertainty, consistently led to higher appropriateness ratings even in both types of situations. This suggests that the models’ responses were more influenced by the modifiers rather than the situations. The models’ reliance on the semantic information inherent in the modifiers, rather than utilizing the contextual cues, indicates that the models are not effectively applying visually presented contextual information to ignorance inference.

\section{Experiment 2}
\subsection{Prompt}
Experiment 2 was conducted with the assumption that providing multiple pieces of contextual information would bring about pragmatic interpretation. Thus, another contextual cue, QUD, was added to the previous experimental setup. For QUDs, two types of conditions were designed, such as ‘howmany’ and ‘polar’ as below.

In the howmany condition, the question focuses on a specific number of target objects, leading responses containing superlative and comparative modifiers to introduce uncertainty or information gaps, which in turn trigger ignorance inferences. In contrast, the ‘polar’ condition elicits a simple yes/no response, placing minimal demands on numerical precision. Accordingly, it serves as a baseline for assessing the effects of the howmany QUD.

The appropriateness of the text in response to either of the two questions was measured on a 1-7 scale. For each condition, all experimental sets were repeated 5 times, resulting in a total of 4,200 individual responses.

\vspace{1em}
\renewcommand{\arraystretch}{2}  
\begin{tabular}{|p{0.4\textwidth}|}
\hline
Is the following answer to the question appropriate for the given image? \\
Please reply with a single integer between 1 and 7, where 1 means “not at all appropriate” and 7 means “completely appropriate.” \\

\textit{(QUD: howmany)}\\[-12pt]
\textbf{Question:} How many \{objects\} did you find in the boxes?\\[-12pt]
\textbf{Answer:} \{text\} \\

\textit{(QUD: polar)}\\[-12pt]
\textbf{Question:} Did you find four \{objects\} in the boxes?\\[-12pt]
\textbf{Answer:} \{text\} \\
\hline
\end{tabular}
\vspace{1em}

\subsection{Result}
Figure 3 shows the mean scores for the appropriateness of image-text pairs, when the text was given as a response of howmany and polar questions, based on the types of modifiers, situations, and models. The statistical analysis was the same as in experiment 1, with QUD added as an independent variable in the fixed effects. For QUD, the polar condition was set as the reference level.

In the howmany condition, we observed that the score for bare numerals increased compared to the results from Experiment 1. In most cases observed in GPT and Gemini, bare numerals received the highest score, with the order being \textit{bare > superlative > comparative}. For the precise condition of Gemini, the order was \textit{superlative > bare > comparative}, but again, the score for bare numerals increased compared to the previous experiment.

Statistical analysis revealed that, for both GPT and Gemini, no significant effects were observed for the modifier (GPT: \textit{p} = 0.11, < 0.001 | Gemini: \textit{p} < 0.001, = 0.46). However, main effects were captured for the two contextual cues, situation (GPT: \textit{p} < 0.05 | Gemini: \textit{p} < 0.001) and QUD (GPT: \textit{p} < 0.001 | Gemini: \textit{p} < 0.001). Additionally, significant interactions were observed between each contextual cue and the modifier, specifically for the interactions of situation and modifier (GPT: \textit{p} < 0.05, 0.001 | Gemini: \textit{p} < 0.001), and QUD and modifier (GPT: \textit{p} < 0.001, 0.05 | Gemini: \textit{p} < 0.001). However, the interaction between the two contextual cues, situation and QUD (GPT: \textit{p} = 0.58 | Gemini: \textit{p} = 0.92), as well as the interaction of modifier, situation, and QUD (GPT: \textit{p} = 0.69, 0.92 | Gemini: \textit{p} < 0.05, = 0.92), were not significant. This finding suggests that while the influence of modifiers remains present, the increased availability of contextual information appears to guide the models toward a more context-driven interpretation strategy.
 
On the other hand, in case of Claude, the scores followed the order of \textit{bare > superlative > comparative} in the precise situation, while the order was \textit{superlative > bare > comparative} in the approximate situation. Although this does not perfectly align with human experimental results, it reflects a pattern similar to our expectations, where bare numerals would be preferred in the precise situation, and either superlative or comparative modifiers would be preferred in the approximate situation. Furthermore, statistical analysis showed a significant effect only in the interaction of both contextual cues, situation and QUD (\textit{p} < 0.01). Considering that in Experiment 1, Claude did not show a similar pattern to human results based on visually encoded context, and that its results were strongly influenced by the modifiers, and the combination of modifier and situation, these findings suggest the possibility that when multiple contextual cues are provided, the model may combine them in a way that aligns more closely with human ignorance inferences, showing a tendency to rely more on contextual cues than on semantic modifiers.

In the polar as a control condition, the appropriateness score for bare numerals was higher compared to the result of Experiment 1, but still lower than in the howmany condition across all the models.

In summary, when the contextual cue QUD was added to the previous experimental setup, GPT and Gemini showed a tendency to prefer bare numerals, which refer to precise knowledge, compared to when only a single contextual cue was provided. In contrast, Claude demonstrated a more integrated approach by utilizing multiple contextual cues together, leading to an interpretation that was closer to pragmatic inference. Despite differences in how these models interpreted the stimuli, all models showed a common pattern of shifting reliance from modifiers to contextual cues when multiple cues were provided.

\begin{figure}[t]
  \includegraphics[width=\columnwidth]{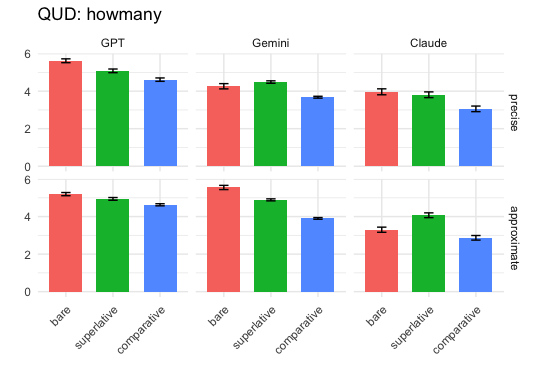}
  \includegraphics[width=\columnwidth]{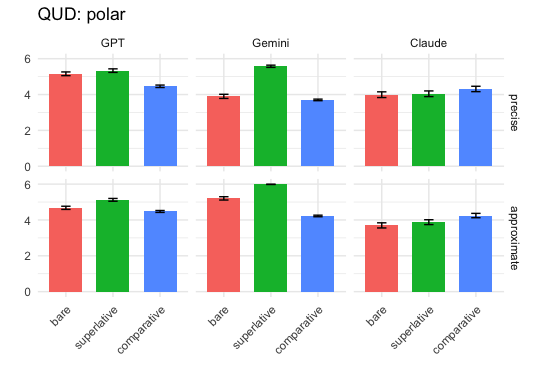}
  \caption{Result of Experiment2 — Mean scores for the appropriateness of image-text pairs based on modifiers, situations, and models across QUDs}
\end{figure}

\begin{figure}[t]
  \includegraphics[width=\columnwidth]{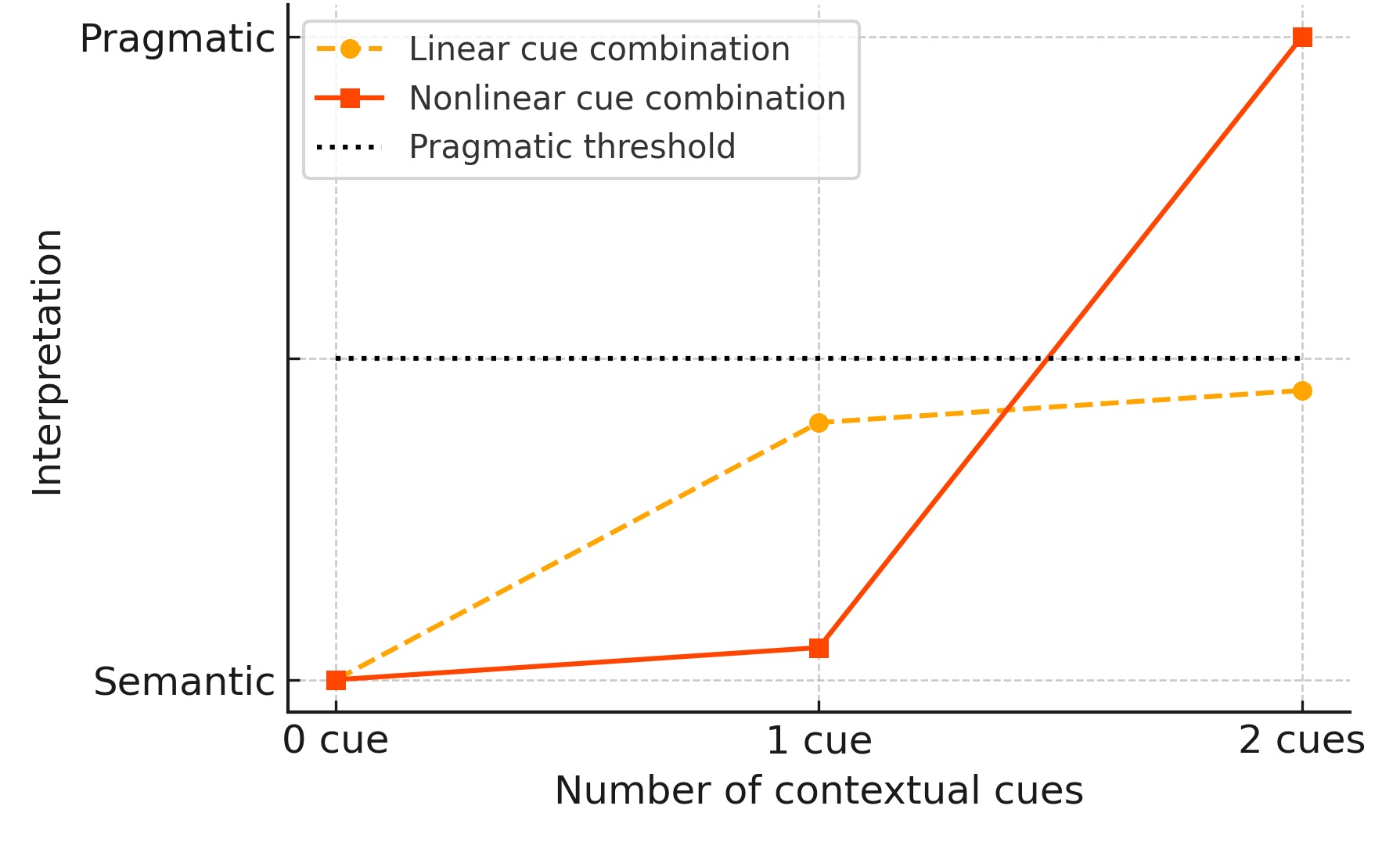}
  \caption{Modeling a threshold effect via linear and nonlinear cue combination as a function of contextual cue number (adapted from \citealp{parker2019cue})}
\end{figure}

\section{Discussion}
This study aimed to investigate the influence of contextual cues in the interpretation of ignorance inference within VLMs. In Experiment 1, we investigated how visually depicted situation (precise and approximate) and different types of modifiers (bare, superlative, and comparative) influenced appropriateness ratings of image-text pairs. Results revealed that appropriateness ratings consistently followed the order of \textit{superlative > comparative > bare} across almost all models, regardless of situation types. This pattern suggests that the models primarily relied on the semantic features of modifiers rather than incorporating contextual information into their judgments.

Building upon these findings, Experiment 2 introduced an additional contextual cue, QUD, with two conditions (howmany and polar). This experiment aimed to determine whether multiple contextual cues would facilitate more sophisticated pragmatic inference. Interestingly, when presented with the howmany QUD, both GPT and Gemini models gave higher appropriateness ratings to bare numerals, which aligned with expectations for precise situation, but not for approximate situation. In our analysis, these models showed greater improvement in providing precise information than in engaging in pragmatic reasoning. While the influence of modifiers remained, there was a modest increase in sensitivity to contextual cues.
 
In contrast, Claude demonstrated a more integrated approach to contextual reasoning, using both situation and QUD simultaneously. This integration pattern suggests Claude may be moving closer to human-like pragmatic reasoning, which typically involves holistic consideration of multiple contextual factors. This leads to the assumption that Claude may have benefited from cue combination, where the presence of two contextual cues, rather than a single cue, led to pragmatic interpretation. 

This pattern resonates with \citet{parker2019cue}'s cue combination scheme, which posits the processing benefit of retrieval cues for anaphora in memory is not merely additive but emerges nonlinearly when multiple cues are jointly available. Extending this idea to our findings, contextual cue combination for ignorance inference in VLMs may similarly follow the nonlinear cue combination method: one contextual cue alone may not significantly affect the context-sensitive reasoning, but the addition of the second cue increases the “cue weight” enough to reach the threshold for pragmatic interpretation. This threshold effect is visualized in Figure 4, which contrasts linear and nonlinear cue integration patterns as a function of contextual cue number.

Then, do GPT and Gemini follow the linear cue combination? It would be insufficient to characterize these models’ behavior merely as examples of linear cue combination. Rather, the observed pattern suggests a difference in how these models represent and utilize contextual information. While Claude appears to engage in combining two contextual cues into a unified pragmatic representation, GPT and Gemini exhibit a pattern of local alignment, in which modifiers are evaluated separately with each cue, but the cues themselves remain structurally unbound. In this sense, their responses are not limited because they combine cues linearly, but because their internal processing architecture does not support contextual cue combination in the first place. Consequently, their outputs reflect a tendency to prioritize informational precision over pragmatic reasoning. As more cues become available, the models tend to converge on more semantically determinate interpretations, aiming to reduce uncertainty in a localized manner rather than resolving it through holistic, context-sensitive inference.

Taken together, these findings offer new insights into how current VLMs differ in their capacity for contextual cue combination in pragmatic inference. By introducing multiple types of cues—both visual and linguistic—within a controlled experimental setting, this study provides empirical evidence that not all models process contextual information in the same way, and that the ability to integrate multiple cues holistically may serve as a crucial indicator of emerging pragmatic competence in VLMs. In doing so, this research contributes to the growing body of work on multimodal language processing by highlighting the need to evaluate not only what models generate and understand, but also how they integrate diverse contextual cues to infer meaning.

\section{Conclusion}

This study examined whether and how current VLMs engage in pragmatic inference, particularly focusing on ignorance implicatures, when provided with visual and linguistic contextual cues. Through two experiments manipulating modifier types and contextual cues—including situation and QUDs—we found that not all VLMs process such information in the same way. Claude demonstrated the ability to integrate multiple contextual cues into a unified interpretation, exhibiting a threshold effect in pragmatic reasoning when both contextual cues were available. In contrast, GPT and Gemini tended to treat these cues independently, prioritizing precision over context-sensitive inference. This suggests a fundamental difference not only in cue weighting tendencies but also in how models internally represent and combine contextual information.

By systematically evaluating ignorance implicatures across VLMs, this study contributes to our understanding of the mechanisms underlying pragmatic behavior in VLMs. Importantly, it highlights that the capacity for contextual cue combination may serve as one of the key indicators of emerging pragmatic competence in VLMs. These findings open new directions for evaluating and developing VLMs that move beyond literal interpretation toward more human-like pragmatic inference.

\section*{Acknowledgments}
This research was supported by Brian Impact Foundation, a non-profit organization dedicated to the advancement of science and technology for all.
\nocite{*}
\bibliography{latex/custom}

\section*{Limitations}
This study has several limitations that offer directions for future research. First, our experiments focused on a specific pragmatic phenomenon involving modified numerals, which allowed for a controlled testbed but may limit the generalizability of the findings. Extending the investigation to other types of pragmatic inferences would provide a broader understanding of VLMs’ pragmatic competence. Second, although we tested three state-of-the-art models—GPT-4o, Gemini 1.5 Pro, and Claude 3.5—the results may not fully generalize to other architectures, including open-source models with different training paradigms. Expanding the model pool would help assess the robustness of cue integration effects. Lastly, while our study builds on prior human experiments, it does not include a direct comparison with human performance under identical conditions. Such empirical comparisons would clarify whether model behavior reflects genuine pragmatic reasoning or merely statistical alignment with training data.
\\

\appendix
\section{Appendix}
\label{sec:appendix}

\noindent
\begin{minipage}{\columnwidth}
  \centering
  \resizebox{\columnwidth}{!}{%
  \begin{tabular}{lcccc}
    \toprule
    \textbf{} & \textbf{Estimate} & \textbf{Std} & \textbf{\textit{t}} & \textbf{\textit{p}-value} \\
    \midrule
    (Intercept) & 4.24 & 0.17 & 24.31 & \textbf{<0.001} \\
    Situation & -0.09 & 0.21 & -0.43 & 0.66 \\
    Modifier - Superlative & 1.51 & 0.13 & 11.1 & \textbf{<0.001} \\
    Modifier - Comparative & 1.01 & 0.13 & 7.37 & \textbf{<0.001} \\
    Situation:Modifier - Superlative & -0.17 & 0.09 & -1.81 & 0.06 \\
    Situation:Modifier - Comparative & -0.09 & 0.09 & -0.96 & 0.33 \\
    \bottomrule
  \end{tabular}}
  \captionof{table}{Summary of fixed effects from mixed-effects logistic regression models by GPT-4o in Experiment 1}
  \label{tab:stat_gpt_exp1}
\end{minipage}

\noindent
\begin{minipage}{\columnwidth}
  \centering
  \resizebox{\columnwidth}{!}{%
  \begin{tabular}{lcccc}
    \toprule
    \textbf{} & \textbf{Estimate} & \textbf{Std} & \textbf{\textit{t}} & \textbf{\textit{p}-value} \\
    \midrule
    (Intercept) & 2.18 & 0.11 & 20.51 & \textbf{<0.001} \\
    Situation & 0.43 & 0.13 & 3.24 & \textbf{<0.01} \\
    Modifier - Superlative & 0.52 & 0.07 & 7.12 & \textbf{<0.001} \\
    Modifier - Comparative & 0.47 & 0.07 & 6.32 & \textbf{<0.001} \\
    Situation:Modifier - Superlative & 0.07 & 0.04 & 1.61 & 0.11 \\
    Situation:Modifier - Comparative & -0.17 & 0.04 & -3.67 & \textbf{<0.001} \\
    \bottomrule
  \end{tabular}}
  \captionof{table}{Summary of fixed effects from mixed-effects logistic regression models by Gemini 1.5 Pro in Experiment 1}
  \label{tab:stat_gemini_exp1}
\end{minipage}

\noindent
\begin{minipage}{\columnwidth}
  \centering
  \resizebox{\columnwidth}{!}{%
  \begin{tabular}{lcccc}
    \toprule
    \textbf{} & \textbf{Estimate} & \textbf{Std} & \textbf{\textit{t}} & \textbf{\textit{p}-value} \\
    \midrule
    (Intercept) & 2.74 & 0.26 & 10.21 & \textbf{<0.001} \\
    Situation & -0.01 & 0.33 & -0.04 & 0.96 \\
    Modifier - Superlative & 0.61 & 0.19 & 3.19 & \textbf{<0.01} \\
    Modifier - Comparative & 0.42 & 0.19 & 2.22 & \textbf{<0.05} \\
    Situation:Modifier - Superlative & -0.33 & 0.09 & -3.71 & \textbf{<0.001} \\
    Situation:Modifier - Comparative & -0.78 & 0.09 & -8.64 & \textbf{<0.001} \\
    \bottomrule
  \end{tabular}}
  \captionof{table}{Summary of fixed effects from mixed-effects logistic regression models by Claude 3.5 in Experiment 1}
  \label{tab:stat_claude_exp1}
\end{minipage}

\noindent
\begin{minipage}{\columnwidth}
  \centering
  \resizebox{\columnwidth}{!}{%
  \begin{tabular}{lcccc}
    \toprule
    \textbf{} & \textbf{Estimate} & \textbf{Std} & \textbf{\textit{t}} & \textbf{\textit{p}-value} \\
    \midrule
    (Intercept) & 5.15 & 0.15 & 33.56 & \textbf{<0.001} \\
    Situation & -0.48 & 0.20 & -2.39 & \textbf{<0.05} \\
    QUD & 0.46 & 0.07 & 6.24 & \textbf{<0.001} \\
    Modifier - Superlative & 0.17 & 0.11 & 1.59 & 0.11 \\
    Modifier - Comparative & -0.69 & 0.11 & -6.26 & \textbf{<0.001} \\
    Situation:Modifier - Superlative & 0.26 & 0.10 & 2.56 & \textbf{<0.05} \\
    Situation:Modifier - Comparative & 0.49 & 0.10 & 4.71 & \textbf{<0.001} \\
    QUD:Modifier - Superlative & -0.71 & 0.10 & -6.81 & \textbf{<0.001} \\
    QUD:Modifier - Comparative & 0.30 & 0.10 & -2.94 & \textbf{<0.05} \\
    Situation:QUD & 0.05 & 0.10 & 0.54 & 0.58 \\
    Situation:QUD:Modifier - Superlative & 0.01 & 0.14 & -0.38 & 0.69 \\
    Situation:QUD:Modifier - Comparative & -0.05 & 0.14 & 0.09 & 0.92 \\
    \bottomrule
  \end{tabular}}
  \captionof{table}{Summary of fixed effects from mixed-effects logistic regression models by GPT-4o in Experiment 2}
  \label{tab:stat_gpt_exp2}
\end{minipage}

\noindent
\begin{minipage}{\columnwidth}
  \centering
  \resizebox{\columnwidth}{!}{%
  \begin{tabular}{lcccc}
    \toprule
    \textbf{} & \textbf{Estimate} & \textbf{Std} & \textbf{\textit{t}} & \textbf{\textit{p}-value} \\
    \midrule
    (Intercept) & 3.83 & 1.45 & 26.27 & \textbf{<0.001} \\
    Situation & 1.32 & 1.19 & 11.12 & \textbf{<0.001} \\
    QUD & 0.35 & 6.22 & 5.72 & \textbf{<0.001} \\
    Modifier - Superlative & 1.74 & 1.78 & 9.75 & \textbf{<0.001} \\
    Modifier - Comparative & 0.13 & 1.78 & -0.73 & 0.46 \\
    Situation:Modifier - Superlative & -0.92 & 8.66 & -9.29 & \textbf{<0.001} \\
    Situation:Modifier - Comparative & -0.81 & 8.66 & -10.61 & \textbf{<0.001} \\
    QUD:Modifier - Superlative & -1.45 & 8.69 & -16.66 & \textbf{<0.001} \\
    QUD:Modifier - Comparative & -0.39 & 8.69 & -4.44 & \textbf{<0.001} \\
    Situation:QUD & 0.01 & 0.09 & 0.10 & 0.92 \\
    Situation:QUD:Modifier - Superlative & -0.30 & 0.12 & -2.43 & \textbf{<0.05} \\
    Situation:QUD:Modifier - Comparative & -0.01 & 0.12 & -0.09 & 0.92 \\
    \bottomrule
  \end{tabular}}
  \captionof{table}{Summary of fixed effects from mixed-effects logistic regression models by Gemini 1.5 Pro in Experiment 2}
  \label{tab:stat_gemini_exp2}
\end{minipage}

\noindent
\begin{minipage}{\columnwidth}
  \centering
  \resizebox{\columnwidth}{!}{%
  \begin{tabular}{lcccc}
    \toprule
    \textbf{} & \textbf{Estimate} & \textbf{Std} & \textbf{\textit{t}} & \textbf{\textit{p}-value} \\
    \midrule
    (Intercept) & 3.99 & 0.29 & 13.55 & \textbf{<0.001} \\
    Situation & -0.29 & 0.40 & -0.73 & 0.46 \\
    QUD & -0.02 & 0.08 & -0.29 & 0.76 \\
    Modifier - Superlative & 0.05 & 0.13 & 0.37 & 0.70 \\
    Modifier - Comparative & 0.32 & 0.13 & 2.34 & \textbf{<0.05} \\
    Situation:Modifier - Superlative & 0.12 & 0.12 & 1.03 & 0.29 \\
    Situation:Modifier - Comparative & 0.22 & 0.12 & 1.84 & 0.06 \\
    QUD:Modifier - Superlative & -0.20 & 0.12 & -1.68 & 0.09 \\
    QUD:Modifier - Comparative & -1.22 & 0.12 & -9.92 & \textbf{<0.001} \\
    Situation:QUD & -0.37 & 0.12 & -2.99 & \textbf{<0.01} \\
    Situation:QUD:Modifier - Superlative & 0.80 & 0.17 & 1.41 & 0.15 \\
    Situation:QUD:Modifier - Comparative & 0.24 & 0.17 & 4.56 & \textbf{<0.001} \\
    \bottomrule
  \end{tabular}}
  \captionof{table}{Summary of fixed effects from mixed-effects logistic regression models by Claude 3.5 in Experiment 2}
  \label{tab:stat_claude_exp2}
\end{minipage}

\bibliographystyle{acl_natbib}

\end{document}